\documentclass[10pt,twocolumn,letterpaper]{article}

\usepackage{cvpr}

\usepackage{microtype}
\usepackage{times}
\usepackage{epsfig}
\usepackage{subfigure}
\usepackage{graphicx}
\usepackage{amsmath}
\usepackage{amssymb}
\usepackage{setspace}
\usepackage{booktabs}       

\DeclareMathOperator{\E}{\mathbb{E}}

\usepackage[pagebackref=true,breaklinks=true,letterpaper=true,colorlinks,bookmarks=false]{hyperref}

\cvprfinalcopy 

\def\cvprPaperID{3662} 

\ifcvprfinal\fi
\begin{document}

\title{Effects of Loss Functions And \\ Target Representations on Adversarial Robustness}

\author{Sean Saito\\
SAP Asia, Singapore\\
{\tt\small sean.saito@sap.com}
\and
Sujoy Roy\\
SAP Asia, Singapore\\
{\tt\small sujoy.roy@sap.com}
}

\maketitle

\begin{abstract}

Understanding and evaluating the robustness of neural networks under adversarial settings is a subject of growing interest. Attacks proposed in the literature usually work with models trained to minimize cross-entropy loss and output softmax probabilities. In this work, we present interesting experimental results that suggest the importance of considering other loss functions and target representations, specifically, (1) training on mean-squared error and (2) representing targets as codewords generated from a random codebook. We evaluate the robustness of neural networks that implement these proposed modifications using existing attacks, showing an increase in accuracy against untargeted attacks of up to 98.7\% and a decrease of targeted attack success rates of up to 99.8\%. Our model demonstrates more robustness compared to its conventional counterpart even against attacks that are tailored to our modifications. Furthermore, we find that the parameters of our modified model have significantly smaller Lipschitz bounds, an important measure correlated with a model's sensitivity to adversarial perturbations.

\end{abstract}

\section{Introduction}

Neural networks produce state-of-the-art results across a large number of domains (\cite{krizhevsky2009learning}, \cite{vaswani2017attention}, \cite{van2016wavenet}, \cite{hannun2014deep}). Despite increasing adoption of neural networks in commercial settings, recent work has shown that such algorithms are susceptible to inputs with imperceptible perturbations meant to cause misclassification (\cite{szegedy2013intriguing}, \cite{goodfellow2014explaining}). It is thus important to investigate additional vulnerabilities as well as defenses against them.


\begin{figure}[ht]
\centering
\includegraphics[width=0.94\linewidth]{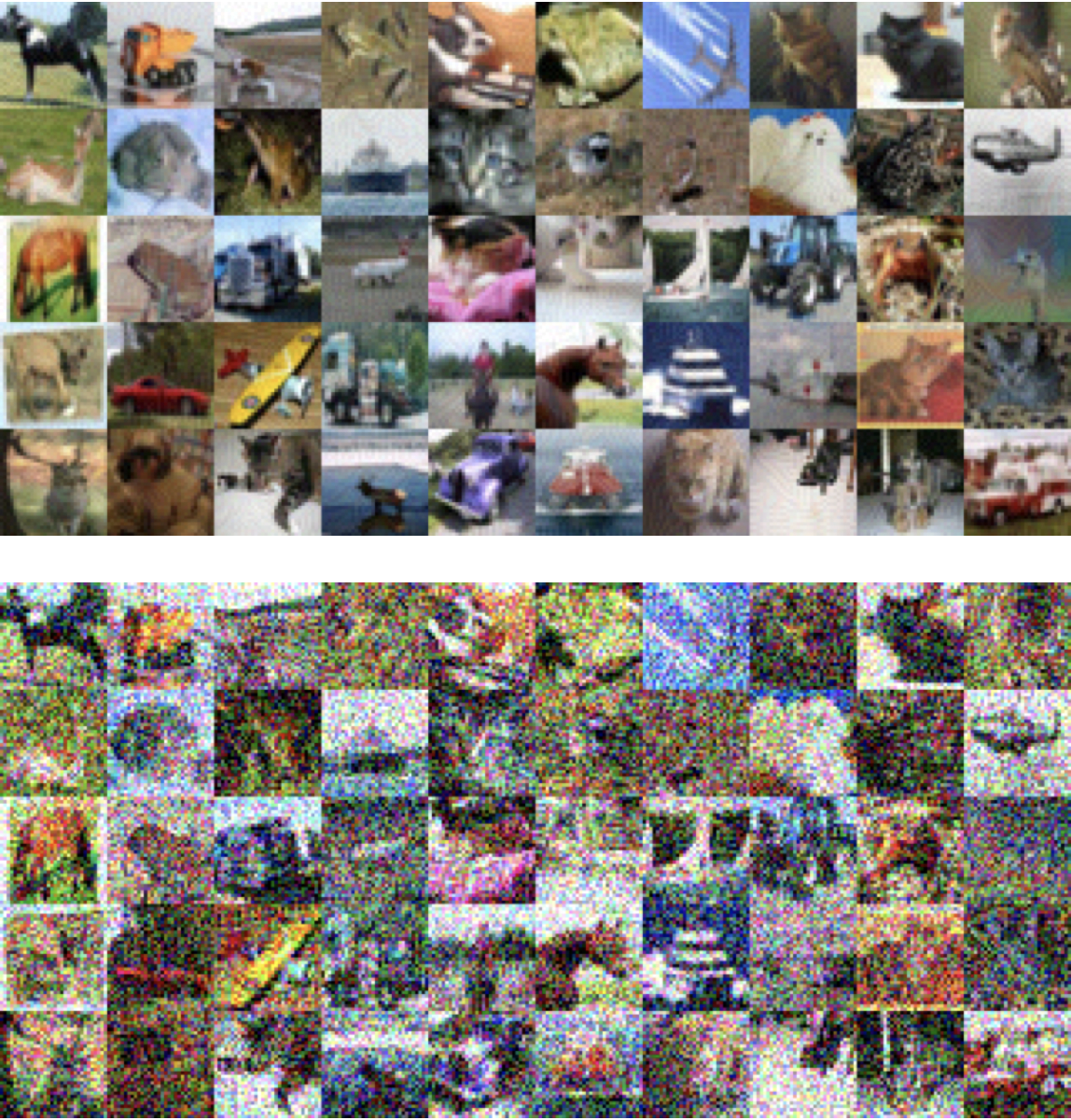}
\caption{\textbf{Top.} Adversarial images generated by the \textbf{targeted white-box Madry et al. attack} using the CIFAR-10 test dataset and a DenseNet model with one-hot target representations and softmax outputs trained to minimize cross-entropy. \textbf{Bottom.} Adversarial images generated under the identical setting, but using a DenseNet model trained to minimize the mean-squared error between tanh outputs and codeword targets. We choose the smallest $\epsilon$ which achieve success rates of 100\% against each model (\textbf{0.03} and \textbf{0.35}, respectively).}

\label{fig:fixed_success_rate_small}
\end{figure}

In this paper we investigate the problem of adversarial attacks on image classification systems. Attacks so far have only considered the conventional neural network architecture which outputs softmax predictions and is trained by minimizing the cross-entropy loss function. We thus propose and evaluate the robustness of neural networks against adversarial attacks with the following modifications:

\begin{itemize}
    \item Train the model to minimize mean-squared error (MSE), rather than cross-entropy.
    \item Replace traditional one-hot target representations with codewords generated from a random codebook.
\end{itemize}

We evaluate our proposed modifications from multiple angles. First, we measure the robustness of the modified model using attacks under multiple threat scenarios. Secondly, we introduce an attack which, without sacrificing its efficacy towards conventional architectures, is tailored to our proposed modifications. 
Finally, we conduct spectral analysis on the model's parameters to compute their upper Lipschitz bounds, a measure that has been shown to be correlated with a model's robustness. Our results in Section \ref{sec:experimental_results} demonstrate that, across all three evaluations, our proposed model displays increased robustness compared to its conventional counterpart.

\section{Background}

\subsection{Neural networks} 

A neural network is a non-linear function $F_\theta$ that maps data $x \in \mathbb{R}^n$ to targets $y \in \mathbb{R}^d$, where $n, d$ are the dimensions of the input and target spaces, respectively, and $\theta$ represents the parameters of the neural network. For conventional neural networks and classification tasks, $y$ is typically a one-hot representation of the class label and $d$ is the number of classes in the dataset. In this work, we use the DenseNet architecture \cite{huang2017densely} as the existing benchmark, which has recently produced state-of-the-art results on several image datasets. 


\subsection{Adversarial examples} 
The goal of an adversarial attack is to cause some misclassification from the target neural network. In particular, \cite{szegedy2013intriguing} has shown that it is possible to construct some $\Tilde{x}$ by adding minimal perturbations to the original input $x$ such that the model misclassifies $\Tilde{x}$. Here, $\Tilde{x}$ is commonly referred to as an \textit{adversarial example}, while the original data $x$ is referred to as a \textit{clean example}. Apart from image classification, adversarial attacks have been proposed in both natural language and audio domains (\cite{carlini2018audio}, \cite{alzantot2018generating}, \cite{yakura2018robust}).
\subsection{Attacks}
\label{subsec:attacks}

\paragraph{Settings.}We explore two adversarial settings, namely white-box and black-box scenarios. In the white-box setting, the attacker has access to and utilizes the model's parameters, outputs, target representations, and loss function to generate adversarial examples. In the black-box scenario, the attacker has no access to the model's parameters or specifications and only has the ability to query it for predictions. In this work, we employ transfer attacks, a type of black-box attack where adversarial examples are generated using a proxy model which the adversary has access to.

\paragraph{Types.}There are mainly two types of attacks. In a \textit{targeted attack}, an adversary generates an adversarial example so that the target model returns some target class $t$. A targeted attack is evaluated by its \textit{success rate}, which is the proportion of images for which the target class was successfully predicted (the lower the better from the perspective of the defense). On the other hand, in an \textit{untargeted attack}, the attacker causes the model to simply return some prediction $y' \neq y$. It is evaluated by the \textit{accuracy} of the target model, which denotes the proportion of images which failed to get misclassified (the higher the better from the perspective of the defense). 

The following sections describe the attacks used in this work.


\paragraph{Fast Gradient Sign Method (FGSM).} The Fast Gradient Sign Method \cite{goodfellow2014explaining}, one of the earliest gradient-based attacks, generates adversarial examples via:

\vspace{-2mm}
\begin{equation*}
    \Tilde{x} = x + \epsilon \cdot sign(\nabla_x J(x, y_t)) 
\end{equation*}
\vspace{-3mm}

where $J$ is the loss function of the neural network, $y_t$ is the target class, and $\epsilon$ is a parameter which controls the magnitude of the perturbations made to the original input $x$. The gradient, which is taken w.r.t the input, determines which direction each pixel should be perturbed in order to maximize the loss function and cause a misclassification.

\paragraph{Basic Iterative Method (BIM).} The Basic Iterative Method, proposed by \cite{kurakin2016physicalworld}, applies FGSM iteratively to find more effective adversarial examples. 

\paragraph{Momentum Iterative Method (MIM).} The Momentum Iterative Method \cite{dong2018boosting} combines iterative gradient-based attacks with the accumulation of a velocity vector based on the gradient of the loss function.

\paragraph{L-BFGS Attack.} \cite{szegedy2013intriguing} proposed the L-BFGS attack, the first targeted white-box attack on convolutional neural networks, which solves the following constrained optimization problem:

\vspace{-3mm}
\begin{equation*}
\begin{aligned}
& {\text{minimize}}
& & c \cdot \left \| x - \Tilde{x} \right \|^2_2 + L_{F, t}(\Tilde{x})\\
& \text{s.t.}
& & \Tilde{x} \in [0, 1]^n
\end{aligned}
\end{equation*}
\vspace{-2mm}

The above formulation aims to minimize two objectives; the left term measures the distance ($L_2$ norm) between the input and the adversarial example, while the right term represents the cross-entropy loss. It is used only as a targeted attack.

\paragraph{Deep Fool.} The Deep Fool attack, proposed by \cite{moosavi2016deepfool}, is an attack which imagines the decision boundaries of neural networks to be linear hyperplanes and uses an iterative optimization algorithm similar to the Newton-Raphson method to find the smallest $L_2$ perturbation which causes a misclassification. It is used only as an untargeted attack.

\paragraph{Madry et al.} \cite{madry2017towards} proposed an attack based on projected gradient descent (PGD), which relies on local first order information of the target model. The method is similar to FGSM and BIM, except that it uses random starting positions for generating adversarial examples.

\paragraph{Carlini \& Wagner L2 (CWL2)}. The Carlini \& Wagner L2 attack \cite{carlini2016towards} follows an optimization problem similar to that of L-BFGS but replaces cross-entropy with a cost function that depends on the pre-softmax logits of the network. In particular, the attack solves the following problem:


\vspace{-2mm}
\begin{equation*}
\begin{aligned}
    & {\text{minimize}}
    & & \left \| \delta \right \|_2 + c \cdot f(x + \delta) \\
    & \text{s.t.}
    & & x + \delta \in [0, 1]^n \\
\end{aligned}
\end{equation*}
\vspace{-2mm}

where $\delta$ is the perturbation made to the input and $f$ is the objective function:

\vspace{-2mm}
$$f(\Tilde{x}) = \max (\max_{i\neq t}(z(\Tilde{x})_i - z(\Tilde{x})_{t}), 0)$$
\vspace{-2mm}

Here, $z(x)$ represents the pre-softmax logits of the network. In short, the attack aims to maximize the logit value of the target class while minimizing the $L_2$ norm of the input perturbations.

\section{Improving adversarial robustness}

In this work we have two proposals. First, we propose changes to the conventional neural network architecture and target representations to defend against adversarial attacks described in Section \ref{subsec:attacks}. Second, we propose a modified, more effective CWL2 attack that is specifically tailored to our proposed defense.

\subsection{Training on mean-squared error}
\label{subsec:mseloss}

Instead of the conventional cross-entropy loss, we propose to use MSE to compute the error between the output of the model $F_\theta$ and the target $y \in Y$, where $Y$ is the set of target representations for all classes. During inference, we select the output class for which its target representation $y$ yields the smallest euclidean distance to ${\hat y}$.

\subsection{Randomized target representations}
\label{subsec:randomrep}

Instead of using one-hot encoding as target representations, we represent each target class as a codeword from a random codebook. Specifically, the $n$ target representations corresponding to the $n$ classes are sampled once at the beginning of training from a uniform distribution $U(-1, 1)^d$ based on a secret key. 
To match the representation space of the network output and the targets, the conventional softmax layer is replaced with a tanh activation with $d$ outputs.

\subsection{Modified CWL2 attack}
\label{subsec:modifiedcwl2}

The Carlini \& Wagner L2 attack makes several assumptions about the target network's architecture based on its cost function mentioned in Section \ref{subsec:attacks}, namely that the highest logit value corresponds to the most likely class. However, applying our proposed neural network modifications breaks such assumptions, for the output of the network would be tanh activations and the length of the output would not correspond to the number of classes in the dataset. We thus propose a simple modification to the CWL2 attack where the cost function considers the distance $D$ in some metric space between the logits and the targets:

\vspace{-3mm}
$$f(\Tilde{x}) = \max (D(z(\Tilde{x})_{t}, t) - \min_{i\neq t} D(z(\Tilde{x})_i, i), 0)$$
\vspace{-3mm}

Like with the Carlini \& Wagner L2 attack, $f(\Tilde{x}) = 0$ if and only if the model predicts the target class. Using the change-of-variables formulation utilized in \cite{carlini2016towards} to enforce box constraints on the perturbations, our attack finds some $w$ which optimizes the following objective:

\vspace{-3mm}
$$\min \left \| \frac{1}{2}(\tanh(w) + 1) - x  \right \|^2_2 + c \cdot f(\frac{1}{2} (\tanh(w) + 1) $$

where $c$ is a trade-off constant that controls the importance of the size of perturbations (larger values of $c$ allow for larger distortions). For our experiments, we have defined $D(x, y)$ as the euclidean distance.

\subsection{Lipschitz bounds and robustness}
\label{subsec:lipschitz}

Earlier works have suggested that the sensitivity of neural networks towards adversarial perturbations can be measured with the upper Lipschitz bound of each network layer \cite{szegedy2013intriguing}. Parseval Networks \cite{cisse2017parseval}, for example, have introduced a layer-wise regularization technique for improving robustness by enforcing smaller global Lipschitz bounds. More specifically, \cite{cisse2017parseval} have shown that:

\vspace{-3mm}
$$\E [J_{adv}(F(\Tilde{x},\theta), y, \epsilon)] \leq \E [J(F(x, \theta), y)] + \lambda  \Lambda \epsilon$$

where $J_{adv} = \max_{\Tilde{x}:\left \| \Tilde{x} - x \right \| \leq \epsilon} J(F(\Tilde{x}, \theta), y)$ , and $\lambda, \Lambda$ are the upper Lipschitz bounds of $J$ and $F$, respectively. In other words, the efficacy of an adversarial attack depends on the generalization error of the target model as well as the Lipschitz bounds of its layers. This suggests that smaller Lipschitz bounds indicate a more robust model. For both fully-connected and convolutional layers, this can be measured by calculating their operator norms. The operator norm $\left \| \theta^l \right \|$ of the $l$-th fully-connected layer is simply the largest singular value of the weight matrix. The Lipschitz constant of the $l$-th layer is then:

\vspace{-3mm}
$$\Lambda^l = \left \| \theta^l \right \| \Lambda^{l-1}$$
\vspace{-3mm}

For convolutional kernels, we rely on the formulation in \cite{szegedy2013intriguing}, which involves applying the two-dimensional discrete Fourier Transform to find the largest singular values.

Section \ref{subsec:lipschitz_upper_bounds} presents empirical results which demonstrate that simply changing the loss function from cross-entropy to mean-squared error can yield model parameters with significantly smaller Lipschitz bounds.

\section{Experimental setup}

In this section we describe the evaluation datasets, evaluation models and adversarial image generation process.

\subsection{Datasets}
\label{subsec:dataset}

\textbf{CIFAR-10} \cite{krizhevsky2009learning} is a small image classification dataset with 10 classes. It contains 60,000 thumbnail-size images of dimensions 32x32x3, of which 10,000 images are withheld for testing.

\textbf{MNIST} \cite{lecun-mnisthandwrittendigit-2010} is another image classification dataset containing monochromatic thumbnails (28x28) of handwritten digits. It is comprised of 60,000 training images and 10,000 testing images.

\textbf{Fashion-MNIST} \cite{xiao2017/online} is a relatively new image classification dataset containing thumbnail images of 10 different types of clothing (shoes, shirts, etc.) which acts as a drop-in replacement to MNIST.

\subsection{Models evaluated}
\label{subsec:modelsevaluated}

We use three variants of the DenseNet model to generate adversarial examples:

\begin{itemize}
\label{sec:network_variants}
    \item O:SOFTMAX:CE refers to a DenseNet model with softmax activations trained on cross-entropy loss and one-hot target representations.
    \item O:SOFTMAX:MSE refers to a DenseNet model with softmax activations trained on MSE and one-hot target representations.
    \item R:TANH:MSE refers to a DenseNet model with tanh activations trained on MSE using codeword target representations. We used a codeword length of $d=128$.
\end{itemize}

We have evaluated the robustness of the R:TANH:MSE model with different codeword lengths (64, 256, and 1024) but found no significant discrepancies in the results.

\begin{table}[!ht]
\centering
\caption{Parameters changed for each attack}
\vspace{2mm}
\scalebox{0.9}{
\begin{tabular}{ll}
\toprule
Attack              & Modified Parameter  \\
\midrule
Basic Iterative Method  & epsilon ($\epsilon$)     \\ 
Carlini \& Wagner L2  & initial constant ($c$)    \\ 
Deep Fool  & max iterations ($m$)    \\ 
Fast Gradient Sign Method & epsilon ($\epsilon$) \\
L-BFGS Attack  & initial constant ($c$)     \\
Madry et al.  & epsilon ($\epsilon$)     \\ 
Momentum Iterative Method  & epsilon ($\epsilon$)     \\ 
\bottomrule
\end{tabular}
}
\label{table:attack_params}
\end{table}

\begin{table}[!ht]
  \centering
  \caption{Parameters held constant for each attack}
  \vspace{2mm}
  \scalebox{0.8}{
  \begin{tabular}{lcc}
    \toprule
    Attack & \multicolumn{2}{c}{Parameters} \\
    \midrule
    
    Basic Iterative Method & eps\_iter & nb\_iter                   \\
    \cmidrule(r){2-3} 
    & 0.05     & 10 \\
    \midrule
    
    Carlini \& Wagner L2 & binary\_search\_steps & max\_iterations \\
    \cmidrule(r){2-3} 
    & 5 & 1000 \\
    \midrule

    Deep Fool & nb\_candidate & overshoot \\
    \cmidrule(r){2-3} 
    & 10 & 0.02 \\
    \midrule
    
    L-BFGS & binary\_search\_steps & max\_iterations \\
    \cmidrule(r){2-3} 
    & 5 & 1000 \\
    \midrule
    
    Madry et al. & eps\_iter & nb\_iter \\
    \cmidrule(r){2-3} 
    & 0.01 & 40 \\
    \midrule
    
    Momentum Iterative Method & eps\_iter & nb\_iter  \\
    \cmidrule(r){2-3} 
    & 0.06 & 10 \\

    \bottomrule
  \end{tabular}
  }
\label{table:other_params}
\end{table}

\begin{singlespace}
\begin{table*}[!ht]
    \centering
    \caption{Robustness of each model against each \textbf{untargeted white-box attack}. The table below reports the accuracy (the higher the better) of each model against each attack.}
    \vspace{2mm}
    \scalebox{0.95}{
    \begin{tabular}{lllll|llll|llll}
        \toprule
        & \multicolumn{4}{c}{CWL2 ($c$)} & \multicolumn{4}{c}{ MIM ($\epsilon$)} & \multicolumn{4}{c}{Deep Fool ($m$)} \\
        \cmidrule(r){2-13}
        \multicolumn{1}{c}{Setting} & 0.01 & 0.1 & 1 & 10 & 0.01  & 0.05  & 0.1  & 0.2 & 10 & 20 & 30 & 40\\
        
        \cmidrule(r){1-13}
        & \multicolumn{12}{c}{CIFAR-10} \\
        \midrule
        \multicolumn{1}{c}{O:SOFTMAX:CE}   & 0.022 & 0.022 & 0.022 & 0.022 & 0.682 & 0.043 & 0.041 & 0.041 & 0.159 & 0.049 & 0.034 & 0.031  \\ 
        \multicolumn{1}{c}{O:SOFTMAX:MSE} & 0.078 & 0.044 & 0.039 & 0.039 & 0.838 & 0.595 & 0.509  & 0.467 & 0.112 & 0.069 & 0.065 & 0.061  \\ 
        \multicolumn{1}{c}{R:TANH:MSE}   & \textbf{0.583} & \textbf{0.584} & \textbf{0.586} & \textbf{0.585} & \textbf{0.919} & \textbf{0.701}   & \textbf{0.593} & \textbf{0.536} & \textbf{0.583} & \textbf{0.582} & \textbf{0.582} & \textbf{0.582} \\ 

        \cmidrule(r){1-13}
        & \multicolumn{12}{c}{MNIST} \\
        \midrule
        \multicolumn{1}{c}{O:SOFTMAX:CE}   & 0.008 & 0.008 & 0.008 & 0.008 & 0.994 & 0.661  & 0.012 & 0.007 & 0.009 & 0.008 & 0.008 & 0.008  \\ 
        \multicolumn{1}{c}{O:SOFTMAX:MSE} & 0.897 & 0.182  & 0.123 & 0.118 & \textbf{0.997} & 0.986 & 0.956  & 0.831 & 0.100   & 0.074 & 0.059 & 0.049 \\ 
        \multicolumn{1}{c}{R:TANH:MSE}   & \textbf{0.995} & \textbf{0.995} & \textbf{0.975} & \textbf{0.983} & 0.995 & \textbf{0.995} &  \textbf{0.994} & \textbf{0.973} & \textbf{0.815} & \textbf{0.815} & \textbf{0.815} & \textbf{0.815} \\ 

        \cmidrule(r){1-13}
        & \multicolumn{12}{c}{F-MNIST} \\
        \midrule
        \multicolumn{1}{c}{O:SOFTMAX:CE}   & 0.041 & 0.041 & 0.041 & 0.041 & 0.196 & 0.038 & 0.035  & 0.034 & 0.049 & 0.042  & 0.041  & 0.041   \\ 
        \multicolumn{1}{c}{O:SOFTMAX:MSE} & 0.156 & 0.076 & 0.057 & 0.049 & 0.836 & 0.304 & 0.211  & 0.142 & 0.082 & 0.064 & 0.059 & 0.056 \\ 
        \multicolumn{1}{c}{R:TANH:MSE}   & \textbf{0.946} & \textbf{0.942} & \textbf{0.946} & \textbf{0.945} & \textbf{0.902} & \textbf{0.691} & \textbf{0.574} & \textbf{0.568} & \textbf{0.935} & \textbf{0.935} & \textbf{0.935} & \textbf{0.935} \\ 

        \bottomrule

    \end{tabular}
    }
    
    \vspace{1mm}
    
    \scalebox{0.95}{
    \begin{tabular}{lllll|llll|llll}
        & \multicolumn{4}{c}{BIM ($\epsilon$)} & \multicolumn{4}{c}{FGSM ($\epsilon$)} & \multicolumn{4}{c}{Madry et al. ($\epsilon$)} \\
        \cmidrule(r){2-13}
        \multicolumn{1}{c}{Setting} & 0.01 & 0.05 & 0.1 & 0.2 & 0.01  & 0.05  & 0.1  & 0.2 & 0.02 & 0.04 & 0.08 & 0.1 \\
        
        \cmidrule(r){1-13}
        & \multicolumn{12}{c}{CIFAR-10} \\
        \midrule
        \multicolumn{1}{c}{O:SOFTMAX:CE}   & 0.751 & 0.053 & 0.042 & 0.042   
                                           & 0.743 & 0.291 & 0.193 & 0.139 
                                           & 0.301 & 0.050 & 0.041 & 0.041\\ 
        \multicolumn{1}{c}{O:SOFTMAX:MSE} & 0.807 & 0.424 & 0.240 & 0.174  
                                          & 0.879 & \textbf{0.729} & \textbf{0.666} & \textbf{0.535}
                                          & 0.790 & 0.707 & 0.668 & 0.608\\ 
        \multicolumn{1}{c}{R:TANH:MSE}   & \textbf{0.850} & \textbf{0.634} & \textbf{0.390} &                                 \textbf{0.213} & \textbf{0.923} & 0.699 & 0.604 & 0.451
                                    & \textbf{0.923} & \textbf{0.897} & \textbf{0.877} & \textbf{0.839}\\ 

        \cmidrule(r){1-13}
        & \multicolumn{12}{c}{MNIST} \\
        \midrule
        \multicolumn{1}{c}{O:SOFTMAX:CE}   & 0.994 & 0.628 & 0.015 & 0.008
                                           & 0.994 & 0.949 & 0.654 & 0.227 
                                           & 0.983 & 0.809 & 0.263 & 0.008\\ 
        \multicolumn{1}{c}{O:SOFTMAX:MSE} & \textbf{0.997} & \textbf{0.929} & \textbf{0.490} &                                        \textbf{0.196} & \textbf{0.997} & 0.988 &                                                 \textbf{0.985} & \textbf{0.774} & 0.992 & 0.983 &                                         0.975 & 0.896\\ 
        \multicolumn{1}{c}{R:TANH:MSE}   &  0.995 & 0.882 & 0.429 & 0.196 
                                         &  0.995 & \textbf{0.995} & 0.918 & 0.332 
                                         & \textbf{0.995} & \textbf{0.995} & \textbf{0.995} & \textbf{0.993} \\ 

        \cmidrule(r){1-13}
        & \multicolumn{12}{c}{F-MNIST} \\
        \midrule
        \multicolumn{1}{c}{O:SOFTMAX:CE}   &  0.564 & 0.038 & 0.037 & 0.036 
                                           &  0.659 & 0.321 & 0.225 & 0.147
                                           &  0.046 & 0.036 & 0.033 & 0.029 \\ 
        \multicolumn{1}{c}{O:SOFTMAX:MSE} &  \textbf{0.815} & \textbf{0.296} & \textbf{0.176} &                                       \textbf{0.142} & 0.882 & 0.509 & 0.362 &                                                  \textbf{0.224} & 0.731 & 0.542 & 0.425 & 0.315\\ 
        \multicolumn{1}{c}{R:TANH:MSE} &  0.799 & 0.233 & 0.089 & 0.051
                                        & \textbf{0.905} & \textbf{0.671} & \textbf{0.389} & 0.185 & \textbf{0.901} & \textbf{0.863} & \textbf{0.829} & \textbf{0.802}\\ 

        \bottomrule

    \end{tabular}
    }
    
\label{results:untargeted_attacks}
\end{table*}
\end{singlespace}

\subsection{Generating adversarial examples}
\label{sec:adv_generation}


For each dataset mentioned in Section \ref{subsec:dataset}, we train a model on the training set and generate adversarial examples using the test set. For targeted attacks, we randomly sample a target class for each image in the test set.

We evaluate each model's (listed in Section \ref{subsec:modelsevaluated}) robustness against attacks (listed in Table \ref{table:attack_params}) under the white-box setting. For the R:TANH:MSE model, the attacker has access to the codeword representations. We also evaluate model robustness against transfer attacks, a type of black-box attack where adversarial examples are generated using a proxy model which the adversary has access to. Finally, we further measure the robustness of our proposed model using the modified CWL2 attack.


All experiments are implemented using TensorFlow \cite{tensorflow2015-whitepaper}, a popular framework for building deep learning algorithms.

\subsubsection{Attack parameters}

For a given attack, we generate adversarial examples across a range of values for a particular parameter which controls the magnitude of the perturbations made. Table \ref{table:attack_params} lists the parameters which are modified for each attack, whereas Table \ref{table:other_params} lists the parameters held constant. We use the default values defined in Cleverhans for our constant parameters.

\subsubsection{Adapting attacks to our proposed techniques}

The attacks described in Section \ref{subsec:attacks} are implemented using the Cleverhans library \cite{papernot2017cleverhans}. By default, the attacks assume that the model outputs softmax predictions and that the targets are represented as one-hot vectors. Hence the internal loss function for some attacks (e.g. gradient-based iterative attacks) is predefined as cross-entropy. However, because the cross-entropy loss function is not compatible with the R:TANH:MSE model, we have adapted the library to use mean-squared error when the target model has also been trained on mean-squared error. These adaptations are important in preserving the white-box assumption of each attack.

\section{Experimental observations}
\label{sec:experimental_results}

In this section, we present and analyze the performance of the evaluation models under different attack scenarios: untargeted and targeted attacks (Section \ref{subsec:untargeted_and_targeted_results}), black-box attacks (Section \ref{subsec:black_box_results}), and our modified CWL2 attack (Section \ref{results:modifiedcwl2}). Benchmark performances on the original datasets are presented in Section \ref{subsec:clean_test_acc}.

\subsection{Clean test performance}
\label{subsec:clean_test_acc}

Table \ref{clean-acc} lists the accuracy of each model across each clean test dataset. We observe minimal differences in accuracies across the models, and hence our proposed modifications can maintain state-of-the-art classification performances.

\begin{table}[!ht]
\centering
\caption{Accuracy on each clean test dataset.}
\vspace{1mm}
\scalebox{0.9}{
\begin{tabular}{llll}
\toprule
            & CIFAR-10 & MNIST & F-MNIST \\
\midrule
O:SOFTMAX:CE  & 0.933    & 0.996  & 0.948   \\ 
O:SOFTMAX:MSE & 0.931    & 0.997  & 0.948   \\ 
R:TANH:MSE  & 0.930    & 0.996 & 0.945   \\
\bottomrule
\end{tabular}
}
\label{clean-acc}
\end{table}

\begin{singlespace}
\begin{table*}[!ht]
    \centering
    \caption{Robustness of each model against each \textbf{targeted white-box attack}. The table below reports the success rates (the lower the better) of each attack.}
    \vspace{2mm}
    \scalebox{0.95}{
    \begin{tabular}{lllll|llll|llll}
        \toprule
        & \multicolumn{4}{c}{L-BFGS ($c$)} & \multicolumn{4}{c}{BIM ($\epsilon$)} & \multicolumn{4}{c}{Madry et al. ($\epsilon$)}  \\
        \cmidrule(r){2-13}
        \multicolumn{1}{c}{Setting} & 0.01 & 0.1 & 1 & 10 & 0.1 & 0.2 & 0.3 & 0.4 & 0.04 & 0.06 & 0.08 & 0.1 \\
        
        \cmidrule(r){1-13}
        & \multicolumn{12}{c}{CIFAR-10} \\
        \midrule
        \multicolumn{1}{c}{O:SOFTMAX:CE}  & 1.00  & 1.00  & 1.00  & 1.00  & 0.997 & 1.00   & 1.00   & 1.00 & 0.934 & 0.998 & 1.00   & 1.00 \\ 
        \multicolumn{1}{c}{O:SOFTMAX:MSE}   & 0.667 & 0.864 & 0.955 & 0.994 & 0.461 & 0.624 & 0.658 & 0.664 & 0.266 & \textbf{0.343} & \textbf{0.402} & \textbf{0.441}\\ 
        \multicolumn{1}{c}{R:TANH:MSE}   & \textbf{0.272} & \textbf{0.475} & \textbf{0.554} & \textbf{0.564} & \textbf{0.230} & \textbf{0.337} & \textbf{0.353} & \textbf{0.353} & \textbf{0.242} & 0.345 & 0.426 & 0.467 \\ 

        \cmidrule(r){1-13}
        & \multicolumn{12}{c}{MNIST} \\
        \midrule
        \multicolumn{1}{c}{O:SOFTMAX:CE}    & 1.00   & 1.00   & 1.00   & 1.00 & 0.851 & 0.992 & 0.997 & 0.999 & 0.057 & 0.464 & 0.828 & 0.94 \\ 
        \multicolumn{1}{c}{O:SOFTMAX:MSE}  & \textbf{0.040}  & 0.536 & 0.92  & 0.991 & 0.316 & 0.539 & 0.597 & 0.612 & 0.008 & 0.042 & 0.163 & 0.269\\ 
        \multicolumn{1}{c}{R:TANH:MSE}   & 0.045 & \textbf{0.457} & \textbf{0.72} & \textbf{0.776} & \textbf{0.057} & \textbf{0.129} & \textbf{0.169} & \textbf{0.184} & \textbf{0.007} & \textbf{0.068} & \textbf{0.154} & \textbf{0.245}\\ 

        \cmidrule(r){1-13}
        & \multicolumn{12}{c}{F-MNIST} \\
        \midrule
        \multicolumn{1}{c}{O:SOFTMAX:CE}   & 1.00   & 1.00   & 1.00   & 1.00 & 0.957 & 1.00   & 1.00   & 1.00 & 0.999 & 1.00   & 1.00   & 1.00 \\ 
        \multicolumn{1}{c}{O:SOFTMAX:MSE}  & \textbf{0.571} & 0.87  & 0.97  & 0.992 & \textbf{0.457} & \textbf{0.581} & \textbf{0.600} & \textbf{0.603} & \textbf{0.464} & \textbf{0.589} & \textbf{0.648} & \textbf{0.659}\\ 
        \multicolumn{1}{c}{R:TANH:MSE}  & 0.644 & \textbf{0.808} & \textbf{0.826} & \textbf{0.832} & 0.807 & 0.926 & 0.938 & 0.940 & 0.626 & 0.724 & 0.794 & 0.834\\ 

        \bottomrule

    \end{tabular}
    }
    
    \vspace{1mm}
    \scalebox{0.95}{
    \begin{tabular}{lllll|llll|llll}
        & \multicolumn{4}{c}{CWL2 ($c$)} & \multicolumn{4}{c}{MIM ($\epsilon$)} & \multicolumn{4}{c}{FGSM ($\epsilon$)}  \\
        \cmidrule(r){2-13}
        \multicolumn{1}{c}{Setting} & 0.01 & 0.1 & 1 & 10 & 0.1 & 0.2 & 0.3 & 0.4 & 0.1 & 0.2 & 0.3 & 0.4 \\
        
        \cmidrule(r){1-13}
        & \multicolumn{12}{c}{CIFAR-10} \\
        \midrule
        \multicolumn{1}{c}{O:SOFTMAX:CE} & 1.00 & 1.00 & 1.00 & 1.00  
                                         & 1.00 & 1.00 & 1.00 & 1.00
                                         &  0.445 & 0.316 & 0.231 & 0.182\\ 
        \multicolumn{1}{c}{O:SOFTMAX:MSE} &   0.756 & 0.842 & 0.861 & 0.867 
                                          &  0.351 & 0.433 & 0.459 & 0.468
                                          & 0.046 & 0.061 & \textbf{0.071} & 0.083 \\ 
        \multicolumn{1}{c}{R:TANH:MSE} & \textbf{0.368} & \textbf{0.362} & \textbf{0.361} &                                           \textbf{0.346}  
                                       & \textbf{0.095} & \textbf{0.136} & \textbf{0.137} & \textbf{0.160} 
                                       & \textbf{0.028} & \textbf{0.044} & 0.081 & \textbf{0.082}\\ 

        \cmidrule(r){1-13}
        & \multicolumn{12}{c}{MNIST} \\
        \midrule
        \multicolumn{1}{c}{O:SOFTMAX:CE} & 1.00 & 1.00 & 1.00 & 1.00
                                         &  0.908 & 0.997 & 0.998 & 0.998
                                         & 0.095 & 0.131 & 0.115 & 0.113\\ 
        \multicolumn{1}{c}{O:SOFTMAX:MSE} &  0.176 & 0.592 & 0.677 & 0.669  
                                          &  0.153 & 0.319 & 0.347 & 0.357
                                          &  0.014 & \textbf{0.028} & 0.052 & \textbf{0.067}\\ 
        \multicolumn{1}{c}{R:TANH:MSE} & \textbf{0.006} & \textbf{0.006} & \textbf{0.006} &                                       \textbf{0.002}
                                        & \textbf{0.023} & \textbf{0.040} & \textbf{0.052} & \textbf{0.050} 
                                        & \textbf{0.007} & 0.031 & \textbf{0.047} & \textbf{0.067}\\ 

        \cmidrule(r){1-13}
        & \multicolumn{12}{c}{F-MNIST} \\
        \midrule
        \multicolumn{1}{c}{O:SOFTMAX:CE} & 1.00 & 1.00 & 1.00 & 1.00 
                                         & 1.00 & 1.00 & 1.00 & 1.00 
                                         & 0.213 & 0.158 & 0.122 & 0.109\\ 
        \multicolumn{1}{c}{O:SOFTMAX:MSE} &  0.658 & 0.812 & 0.845 & 0.851 
                                          & 0.394 & 0.396 & 0.394 & 0.382
                                          & 0.073 & 0.085 & 0.102 & 0.107\\ 
        \multicolumn{1}{c}{R:TANH:MSE} & \textbf{0.583} & \textbf{0.592} & \textbf{0.576} &                                       \textbf{0.548} 
                                       & \textbf{0.114} & \textbf{0.140} & \textbf{0.143} & \textbf{0.147}
                                       &\textbf{0.048} & \textbf{0.081} & \textbf{0.088} & \textbf{0.097}\\

        \bottomrule

    \end{tabular}
    }
    \label{results:targeted_attacks}
\end{table*}
\end{singlespace}

\subsection{Untargeted and targeted attacks}
\label{subsec:untargeted_and_targeted_results}

Table \ref{results:untargeted_attacks} lists the accuracies of the models against untargeted white-box attacks. Both O:SOFTMAX:MSE and R:TANH:MSE models demonstrate higher accuracies on the adversarial examples compared to the O:SOFTMAX:CE model; we observe an increase in accuracies of up to 98.7\%. Similar results can be observed in Table \ref{results:targeted_attacks}, where the O:SOFTMAX:MSE and R:TANH:MSE models achieve a consistent decrease in attack success rates of up to 99.8\%.



\subsection{Black box attacks}
\label{subsec:black_box_results}

Table \ref{black-box} shows the accuracies of transfer attacks against the O:SOFTMAX:MSE and R:TANH:MSE models. Our proposed models demonstrate more robustness towards black-box attacks compared to the white-box versions with the same configurations. Though this is expected behavior, it is imperative to evaluate a defense under multiple threat scenarios.


\begin{singlespace}
\begin{table}[!ht]
\centering
\caption{Accuracy on adversarial data generated under the \textbf{black-box setting}. All adversarial examples are generated using the O:SOFTMAX:CE model. Both the O:SOFTMAX:MSE and R:TANH:MSE models show higher accuracy towards black-box attacks compared to untargeted white-box attacks.}
\vspace{2mm}
\scalebox{0.85}{
\begin{tabular}{llll}
\toprule
Setting            & CWL2 & Deep Fool & MIM \\
Parameter           & $c=0.01$ & $m=10$ & $\epsilon=0.01$ \\
\midrule
& \multicolumn{3}{c}{CIFAR-10} \\
\midrule
O:SOFTMAX:MSE & 0.483    & 0.451 & 0.895   \\ 
R:TANH:MSE  & 0.612    & 0.617 & 0.926   \\
\midrule
& \multicolumn{3}{c}{MNIST} \\
\midrule
O:SOFTMAX:MSE & 0.996    & 0.984 & 0.997   \\ 
R:TANH:MSE  & 0.996   & 0.973 & 0.995   \\
\midrule
& \multicolumn{3}{c}{F-MNIST} \\
\midrule
O:SOFTMAX:MSE & 0.937    & 0.933 & 0.839   \\ 
R:TANH:MSE  & 0.952    & 0.946 & 0.935   \\
\bottomrule
\end{tabular}
}
\label{black-box}
\end{table}
\end{singlespace}

\begin{singlespace}
\begin{table}[!ht]
    \centering
    \caption{Success rates of the \textbf{targeted Carlini \& Wagner L2 attack} and \textbf{our tailored attack} on the O:SOFTMAX:CE and R:TANH:MSE models.}    
    \vspace{2mm}
    \scalebox{0.87}{
    \begin{tabular}{lll|ll}
        \toprule
        & \multicolumn{2}{c}{CWL2 ($c$)} & \multicolumn{2}{c}{Ours ($c$)}   \\
        \cmidrule(r){2-5}
        \multicolumn{1}{c}{Setting} & 0.1 & 1.0 & 0.1 & 1.0 \\
        
        \cmidrule(r){1-5}
        & \multicolumn{4}{c}{CIFAR-10} \\
        \midrule
        \multicolumn{1}{c}{O:SOFTMAX:CE} & 1.000 & 1.000 & 1.000 & 1.000  \\ 
        \multicolumn{1}{c}{R:TANH:MSE}   & 0.368 & 0.362 & 0.859 & 0.868 \\ 

        \cmidrule(r){1-5}
        & \multicolumn{4}{c}{MNIST} \\
        \midrule
        \multicolumn{1}{c}{O:SOFTMAX:CE} & 1.000 & 1.000 & 1.000 & 1.000  \\ 
        \multicolumn{1}{c}{R:TANH:MSE} & 0.006 & 0.006  & 0.715 & 0.772\\ 

        \cmidrule(r){1-5}
        & \multicolumn{4}{c}{F-MNIST} \\
        \midrule
        \multicolumn{1}{c}{O:SOFTMAX:CE} & 1.000 & 1.000 & 1.000 & 1.000   \\ 
        \multicolumn{1}{c}{R:TANH:MSE} & 0.583 & 0.592 & 0.798 & 0.829 \\ 

        \bottomrule

    \end{tabular}
    }
\label{results:srl2}
\end{table}
\end{singlespace}

\subsection{Modified CWL2 attack}
\label{results:modifiedcwl2}

Table \ref{results:srl2} compares our proposed attack with the CWL2 attack. The results show that our attack maintains its efficacy against O:SOFTMAX:CE models while significantly increasing its success rate against the R:TANH:MSE model up to \textbf{70.9\%}. We note that increasing the initial constant for our attack yields increased success rates, which is aligned with the intuition that the parameter controls the importance of the attack's success as highlighted in Section \ref{subsec:modifiedcwl2}. We also observe that, despite the increase in the attack's efficacy, the R:TANH:MSE model displays more robustness compared to the O:SOFTMAX:CE model, with a decrease in success rates of up to \textbf{28.5\%}.

\subsection{Distortion vs. performance}
\label{subsec:visualization}

On page 1, Figure \ref{fig:fixed_success_rate_small} displays adversarial images generated from targeted white-box Madry et al. attacks on the O:SOFTMAX:CE and R:TANH:MSE models respectively. We choose the lowest $\epsilon$ for which the attack achieves success rates of 100\%. It is clear that the R:TANH:MSE model requires much larger perturbations for an attack to achieve the same success rates as against the O:SOFTMAX:CE model.

\begin{figure}[!ht]
\centering
    \includegraphics[width=0.8\linewidth]{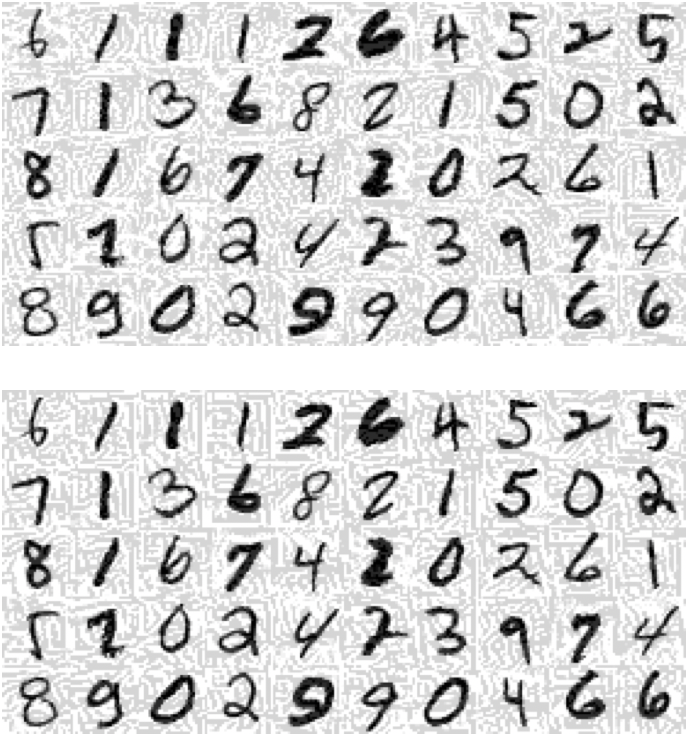}
\caption{\textbf{Top.} Adversarial images generated for MNIST using the \textbf{targeted MIM attack} $(\epsilon=0.1)$ on the O:SOFTMAX:CE model. The attack achieves a success rate of \textbf{90.8\%}. \textbf{Bottom.} Adversarial images generated under the identical setting for the R:TANH:MSE model. The attack achieves a success rate of \textbf{2.3\%}.}
\label{fig:c10_mim_images}
\end{figure}

Figure \ref{fig:c10_mim_images} displays adversarial images generated using the Momentum Iterative Method against both O:SOFTMAX:CE and R:TANH:MSE models where $\epsilon=0.1$. We observe that the R:TANH:MSE model is robust even against adversarial images where the perturbations are clearly perceptible to humans.

\begin{figure}[!th]
\centering
    \includegraphics[width=0.85\linewidth]{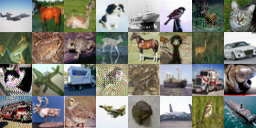}    
    \includegraphics[width=0.85\linewidth]{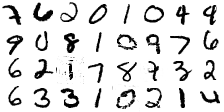}
    \includegraphics[width=0.85\linewidth]{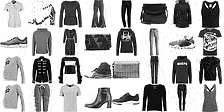}
\caption{Adversarial examples generated by our proposed attack ($c=0.1$) on the R:TANH:MSE model for test images from the CIFAR-10, MNIST, and Fashion-MNIST datasets.}
\label{fig:srl2_images}
\end{figure}

Finally, we visualize adversarial examples generated using our modified CWL2 attack and the R:TANH:MSE model in Figure \ref{fig:srl2_images}, where the attack achieves higher success rates compared to the original attack. The perturbations made to the images are much less perceptible compared to the adversarial examples displayed in Figures \ref{fig:fixed_success_rate_small} and \ref{fig:c10_mim_images}.

\subsection{Comparing upper Lipschitz bounds}
\label{subsec:lipschitz_upper_bounds}

\begin{figure}[!t]
\centering
    \includegraphics[width=0.93\linewidth]{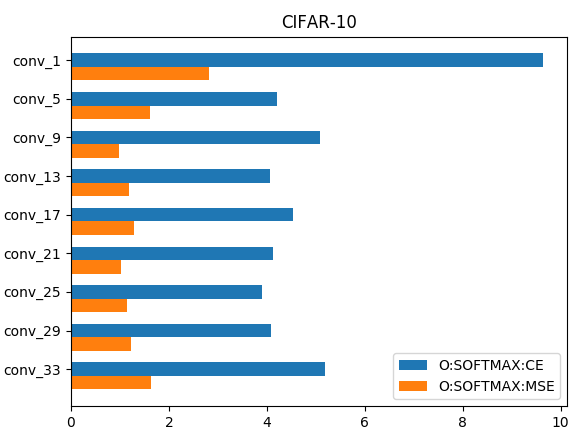}    
    \includegraphics[width=0.93\linewidth]{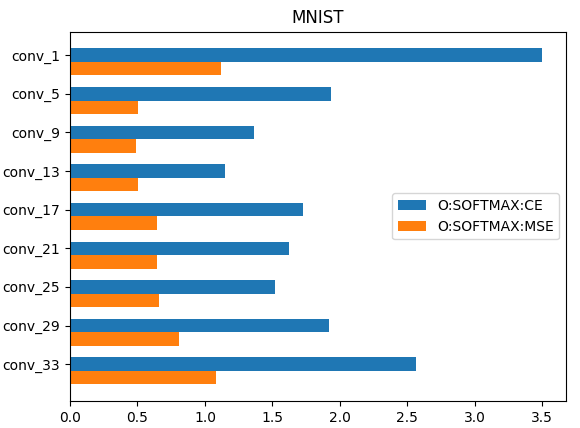}
    \includegraphics[width=0.93\linewidth]{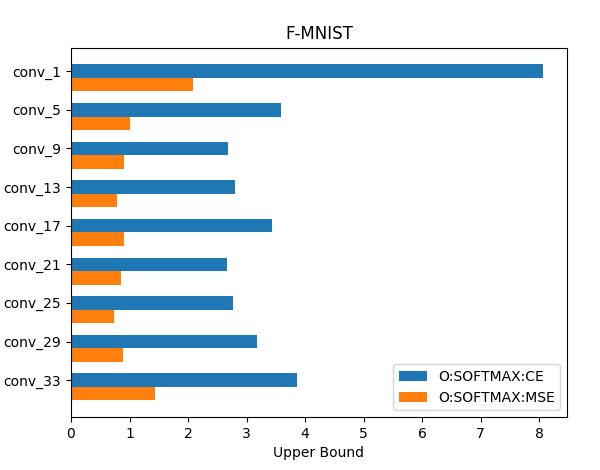}
\caption{Upper Lipschitz bounds of convolutional layers of the O:SOFTMAX:CE and O:SOFTMAX:MSE models for each dataset.}
\label{fig:upper_lipschitz_bounds}
\end{figure}

Figure \ref{fig:upper_lipschitz_bounds} compares the upper Lipschitz bounds of convolutional layers between the O:SOFTMAX:CE and O:SOFTMAX:MSE models. The upper bounds for the O:SOFTMAX:MSE model are consistently smaller than those of the O:SOFTMAX:CE model across each dataset up to a factor of three, supporting our hypothesis that models trained to minimize mean-squared error are more robust to small perturbations.

\section{Related work}
\label{sec:otherelatedwork}

Several defenses have also been proposed. To date, the most effective defense technique is adversarial training (\cite{kurakin2016adversarial}, \cite{wu2018reinforcing}, \cite{sinha2018certifying}, \cite{tramer2017ensemble}), where the model is trained on a mix of clean and adversarial data. This has shown to provide a regularization effect that makes models more robust towards attacks. 

\cite{papernot2015distillation} proposed defensive distillation, a mechanism whereby a model is trained based on soft labels generated by another `teacher' network in order to prevent overfitting. Other methods include introducing randomness to or applying transformations on the input data and/or the layers of the network (\cite{guo2017countering}, \cite{dhillon2018stochastic}, \cite{samangouei2018defense}, \cite{xie2017mitigating}). However, \cite{athalye2018obfuscated} have identified that the apparent robustness of several defenses can be attributed to the introduction of computation and transformations that mask the gradients and thus break existing attacks that rely on gradients to generate adversarial examples. Their work demonstrates that small, tailored modifications to the attacks can circumvent these defenses completely.

\section{Conclusion}
\label{sec:conclusion}

We have reported interesting experimental results demonstrating the adversarial robustness of models that do not follow conventional specifications. We have observed that simply changing the loss function that is minimized during training can greatly impact the robustness of a neural network against adversarial attacks. Our evaluation strategy is manifold, consisting of existing attacks, new attacks adjusted to our proposed modifications, and a spectral analysis of the model's parameters. The increase in robustness observed from experimental results suggests the importance of considering alternatives to conventional design choices when making neural networks more secure. Future work would involve further investigation into the reasons for such modifications to improve the robustness of neural networks.




\end{document}